\RequirePackage{amsmath}
\documentclass{llncs}
\usepackage{graphicx} 
\usepackage{amssymb}
\usepackage{import}
\usepackage[noend]{algpseudocode}
\usepackage{algorithm}
\algblockdefx{MRepeat}{EndRepeat}{\textbf{repeat}}{}
\algnotext{EndRepeat}
\def \figref #1{\figurename\ref{#1}}
\def \tabref #1{\tablename\ref{#1}}
\begin{document}
\mainmatter
\title{Switching Isotropic and Directional Exploration \\with Parameter Space Noise \\in Deep Reinforcement Learning}
\author{
Izumi~Karino\inst{1}\orcidID{0000-0003-3267-3886}\and
Kazutoshi~Tanaka\inst{1}\orcidID{0000-0003-0880-9333}\and
Ryuma~Niiyama\inst{1}\orcidID{0000-0002-9072-8251}\and
Yasuo~Kuniyoshi\inst{1}\orcidID{0000-0001-8443-4161}}
\institute{\textsuperscript{1} The University of Tokyo, Japan,\\
\email{\{karino, tanaka, niiyama, kuniyosh\}@isi.imi.i.u-tokyo.ac.jp}}
\maketitle
\begin{abstract}
This paper proposes an exploration method for deep reinforcement learning based on parameter space noise. Recent studies have experimentally shown that parameter space noise results in better exploration than the commonly used action space noise. Previous methods devised a way to update the diagonal covariance matrix of a noise distribution and did not consider the direction of the noise vector and its correlation. In addition, fast updates of the noise distribution are required to facilitate policy learning. We propose a method that deforms the noise distribution according to the accumulated returns and the noises that have led to the returns. Moreover, this method switches isotropic exploration and directional exploration in parameter space with regard to obtained rewards. We validate our exploration strategy in the OpenAI Gym continuous environments and modified environments with sparse rewards. The proposed method achieves results that are competitive with a previous method at baseline tasks. Moreover, our approach exhibits better performance in sparse reward environments by exploration with the switching strategy.

\end{abstract}
\section{Introduction}
Reinforcement learning (RL) enables an agent to learn an optimal strategy that maximizes the cumulative reward through experience.
Exploration has an important role in the learning process because an agent does not have prior knowledge of an environment.
The optimal action in the agent's knowledge at that time is not necessarily the globally optimal action.
The agent should activate exploration to acquire a better policy. 
By contrast, excess exploration leads to bad performance and does not lead to acquiring better knowledge.

Various strategies have been used to manage this trade-off.
One is the Upper Confidence Bounds (UCB) approach \cite{auer2002finite}.
This method executes exploration by choosing an arm that maximizes the expected reward and that an agent has explored less than the other arms.
The UCB approach in a multi-armed bandit has the benefit that its performance can be evaluated theoretically in terms of regret.
However, using this approach in continuous state-action is difficult.

Exploration in continuous state-action space is conducted using some heuristics.
Intrinsic motivation is one of the approaches. This method activates exploration by adding another reward signal (exploration bonus) to the original task reward. The exploration bonus has various formats. One is the prediction error of the state or action \cite{pathak2017curiosity}. An agent learns a predictor of the forwarding or inverse transition model in parallel with the policy.  Another method considers a novelty of the state using a pseudo-count of visitation with a probability density function \cite{bellemare2016unifying}.
There is also an approach that calculates the exploration bonus as the information gain of the transition model \cite{houthooft2016vime}.
Additionally, there is a style that automatically generates the simplest tasks that are not achievable by a current policy to support learning \cite{sukhbaatar2017intrinsic}.
The intrinsic motivation approach can perform structured exploration even in continuous state-action space.
However, as mentioned in \cite{fortunato2017noisy}, adding an exploration bonus to the original reward can alter the optimal solution to which the agent aims.

Injecting noise is another exploration method. 
Noise can be classified into three types: $\epsilon$-greedy, action space noise and parameter space noise.
$\epsilon$-greedy performs exploration by taking a random action with probability $\epsilon$.
This method is feasible in discrete state-action space, but infeasible in continuous state-action space. 
Action space noise is the most common method in recent deep reinforcement learning.
It perturbs the output of the policy at each step. 
Action space noise can be implemented straightforwardly for most reinforcement learning methods.
Additionally, it is used to implement a stochastic policy. 
However, action space noise is less suited to tasks in which subtle positioning is essential because noise is injected at each step.
Parameter space noise performs exploration by perturbing parameters of the policy.
This noise enables agent consistent exploration because noise is injected at each episode and outputs the same action in the same state, unlike action space noise.
Additionally, the perturbation of policy parameters leads to a complex change of exploratory behavior.
Some researchers have experimentally shown that parameter space noise performs better exploration than action space noise \cite{ruckstiess2008state,plappert2017parameter}.

The evolution strategy (ES) and genetic algorithm (GA) are highly related to parameter space noise. 
Recently, ES and GA have been applied to deep reinforcement learning \cite{salimans2017evolution,such2017deep}.
The features of these methods were analyzed in \cite{zhang2017relationship,lehman2017more}
and they were experimentally shown to be able to acquire a diverse and robust policy.
These methods have the benefit that they can be applied to a broad range of problems for which the Markov decision process (MDP) is less suited.
However, they are less sample efficient than the conventional RL algorithm.
These methods use only the final result, whereas the conventional RL algorithm uses all step information in each episode.

To summarize, parameter space noise has the benefit that it can manage continuous state-action space, does not alter the optimal policy, has the same sample efficiency as the conventional RL approach and performs better exploration than action space noise.
\section{Related Works}
The effectiveness of parameter space noise was first investigated by R{\"u}cksite{\ss} \cite{ruckstiess2008state}.
R{\"u}cksite{\ss} conducted experiments with a policy that had relatively small parameters and specific learning methods: REINFORCE \cite{williams1992simple} and Natural Actor-Critic \cite{peters2005natural}.
Recent research has shown that parameter space noise also works well in deep RL, which includes a large parameter policy \cite{plappert2017parameter,fortunato2017noisy}.

Plappert {\it et al}. formulated how to update a policy with parameter space noise both in an off-policy and on-policy manner \cite{plappert2017parameter}.
Parameter space noise can perform better than $\epsilon$-greedy and action space noise in Deep Q-Networks (DQN) \cite{mnih2015human}, Deep Deterministic Policy Gradient (DDPG) \cite{lillicrap2015continuous}, and Trust Region Policy Optimization \cite{schulman2015trust}.
Plappert {\it et al}. also proposed a method that can connect the variance of parameter space noise to action space noise.
A problem is that this method perturbs a policy by only using isotropic noise. Isotropic noise can be inefficient in large parameter space, such as deep neural networks.
Fortunato {\it et al}. proposed using factorized functional noise in a DQN \cite{mnih2015human} and Asynchronous Advantage Actor-Critic \cite{mnih2016asynchronous}. This approach samples noise from a multivariate normal distribution with a diagonal covariance matrix. An agent learns both the covariance matrix and a policy using gradient method \cite{fortunato2017noisy}.
This method only manages the orthogonal scale of noise distribution and does not consider the correlation between policy parameters.
Additionally, learning a covariance matrix using a gradient method is likely to take a longer time to converge than policy parameters, as mentioned in  \cite{plappert2017parameter}.
Colas {\it et al}. proposed a method that switches the exploration stage and exploitation stage sequentially \cite{colas2018gep}. 
They indicated that this procedure works well in sparse reward and deceptive reward environments.
This method uses a style of parameter space noise in the exploration stage.
This work still has the problem of when to switch the two stages.
The current parameter space noise method has three problems: (1) it considers both the scale and correlation of parameter space noise, (2) it updates covariance quickly for noise to effect policy learning, and 3) it continuously switches the exploration stage and exploitation stage.

To overcome the problems (1) and (2), we combine a part of a method: Policy Improvement with Black Box (PIBB) \cite{stulp2013robot} with Plappert {\it et al}.'s method \cite{plappert2017parameter}. PIBB updates the policy and covariance matrix using one of the ES methods.
We update only the covariance matrix of parameter space noise using this method and update the policy with the commonly used RL method. 
Updating a large dimension policy using ES requires many training steps because ES can only use the overall results of an episode, whereas the common RL approach can use all step information in each episode and use temporal information.
Moreover, we propose a method to switch exploration and exploitative exploration to solve the problem (3).
%
\section{Proposed Method}
\begin{figure}[t]
  \centering
    \includegraphics[width=122mm]{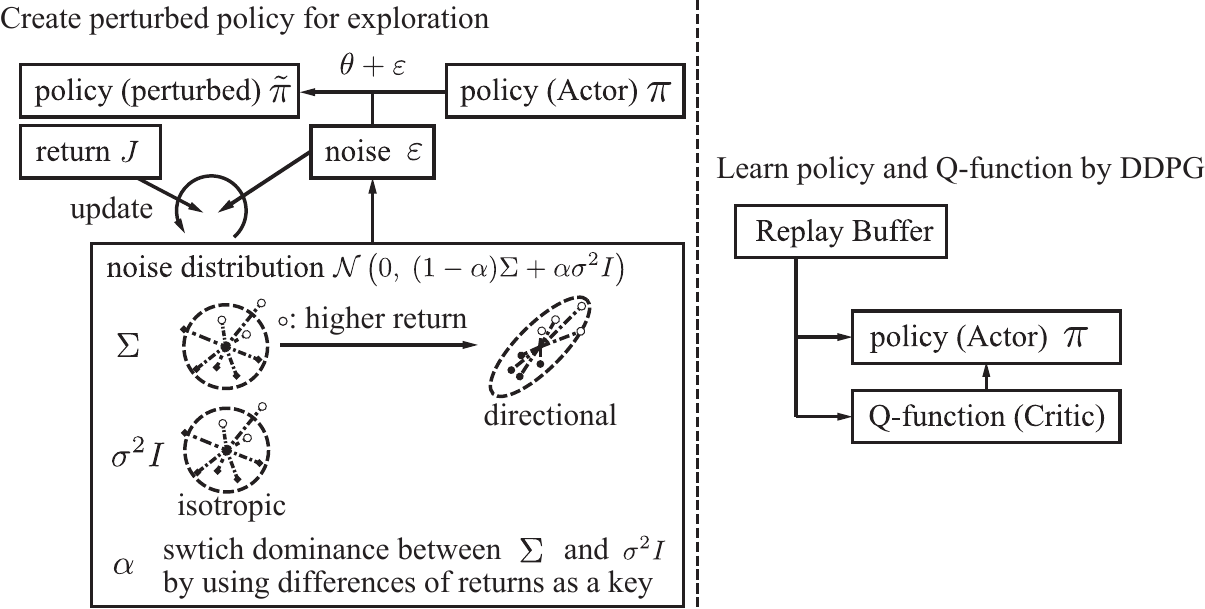}
  \caption{Overview of the learning system}
  \label{fig:overview}
\end{figure}

We consider a standard RL setting. An environment is modeled as an MDP.
The environment consists of state space $\mathcal{S}$, action space $\mathcal{A}$, and reward function $r:\mathcal{S} \times \mathcal{A}\mapsto \mathbb{R}$. An agent learns a policy $\pi:\mathcal{S} \mapsto \mathcal{A}$ through interactions with an environment. 
$\tilde\theta$ and $\tilde\pi$ denote the perturbed policy parameter and perturbed policy parametrized by $\tilde\theta$, respectively.
An overview of the learning system is shown in \figref{fig:overview}.
The update of the policy is conducted by one of the deterministic policy gradient \cite{silver2014deterministic} methods: DDPG \cite{lillicrap2015continuous}.
We propose a method to perturb the policy.

The proposed method samples parameter space noise $\varepsilon$ from the following distribution:
\begin{align}
    \varepsilon \sim \mathcal{N} \left(0, ~(1-\alpha)\Sigma + \alpha \sigma^2 I \right). \label{eq:new_noise}
\end{align}
This method consists of three parts: directional and scalable exploration by covariance matrix $\Sigma \in \mathbb{R}^{N \times N}$, isotropic exploration by a diagonal matrix $\sigma^2 I \in \mathbb{R}^{N \times N}$ and switching the two types of exploration using $\alpha \in \mathbb{R}$. $N$ is the number of policy parameters.
We describe how these modules are interrelated and construct the distribution \eqref{eq:new_noise} in the following subsections.

\subsection{Directional and Scalable Exploration by $\Sigma$}\label{subsec:adaption}
The part of covariance matrix $\Sigma$ in \eqref{eq:new_noise} was designed to solve the two problems (1) and (2) in related works. $\Sigma$ can consider the scale and correlation between policy parameters.
The update of $\Sigma$ is conducted by a part of PIBB \cite{stulp2013robot} based on Covariance Matrix Adaptation\cite{hansen1996adapting}:
\begin{align}
    P_k &= \cfrac{\exp \left(-h \cfrac{J_{\max} - J_k}{J_{\max}-J_{\min}} \right)}
                      {\sum_{k=1}^K \exp \left(-h \cfrac{J_{\max} - J_k}{J_{\max}-J_{\min}} \right)} \label{eq:weight} \\
      \Sigma &\leftarrow \sum_{k=1}^K \left[ P_k (\tilde\theta_k-\theta)(\tilde\theta_k-\theta)^T \right] = \sum_{k=1}^K \left[ P_k \varepsilon_k \varepsilon_k^T \right],
    \label{eq:cov_update}
\end{align}
where variable $h$ is a hyper parameter, $J_k$ is the acquired return in the $k$th episode, $\theta$ denotes the parameters of the policy and $\tilde\theta$ denotes the parameters of the perturbed policy. If $J_{\max}=J_{\min}$, then $P_k=1/K$.
An agent explores an environment with a policy perturbed per episode and obtains rewards. 
Equation \eqref{eq:weight} calculate weights $P_k$ to reflect the goodness of each perturbation $\varepsilon_k$ on the next noise distribution.
Dividing by $J_{\max}-J_{\min}$ enables the use of the same hyper-parameter $h$ in various environments with different scales of the reward function.
The exponential function expands the gap between returns $J_k$. This process makes a weight $P_k$ sensitive to the difference between the returns. 

The scale and correlation of the noise is managed by \eqref{eq:cov_update}. 
Weighting by $P_k$ creates a covariance matrix, which generates more noise with the scale and correlation, which led to higher returns in previous episodes.
This feature can perturb large parameter space efficiently. 
Additionally, it updates the covariance matrix instantly because it does not use a gradient method.
We use small interval $K$ to update $\Sigma$ quickly.
The frequent update of $\Sigma$ can reflect the current policy parameter on current noise.
%
\subsection{Exploration problem of $\Sigma$ for Sparse Reward Environments} \label{subsec:reason}
The update manner \eqref{eq:weight}--\eqref{eq:cov_update} can deform $\Sigma$ adaptively in standard environments.
By contrast, it disables exploration in a sparse reward environment.
A sparse reward environment is a type of environment in which an agent cannot obtain any rewards until it has achieved or almost achieves a task.
In such an environment, exploration is more critical.
An agent with only $\Sigma$ cannot explore a sparse reward environment because the variance of the noise distribution decreases rapidly. 
In a sparse reward environment, an agent generally receives zero returns in the environment. 
Returns with the same value cause the same weights $P_k$ according to \eqref{eq:weight}.
The probability of generating a small noise is high because the noise is generated from a zero-mean Gaussian distribution.
$\Sigma$ is updated by a large population of small noises and the same weights using \eqref{eq:cov_update}.
This process causes a small variance of the next distribution and generates a small noise with higher probability.
By repeating this, the variance of the distribution of parameter space noise decreases sharply.
As a result, the policy is barely perturbed by the noise and the agent cannot explore the environment.
Additionally, when all weights $P_k$ have the same value and a bias emerges in the generated noise, the correlation is accumulated per update, and then an agent can only perturb the policy in a particular direction.
%
\subsection{Isotropic exploration by $\sigma^2 I$}
To solve the variance reduction and fixed correlation of the noise distribution described in \ref{subsec:reason}, isotropic and certain scale exploration is required until an agent obtains rewards. 
A part of the covariance matrix $\sigma^2 I$ performs this type of exploration.
$\sigma$ is updated by Plappert {\it et al}.'s method \cite{plappert2017parameter}:
\begin{align}
  d(\pi, \tilde{\pi}) &= \sqrt{\frac{1}{|A|} \sum_{i=1}^{|A|}\mathbb{E}_s\left[ \left(\pi(s)_i - \tilde{\pi}(s)_i \right)^2 \right]} \label{eq:d}\\
  \sigma &=\begin{cases}
            1.01 \sigma & \mbox{if}~d(\pi,\tilde{\pi}) < \delta \\
            \sigma / 1.01 & \mbox{otherwise}
        \end{cases},
  \label{eq:sigma_plappert}
\end{align}
where $|A|$ is the dimension of actions,
$\delta$ is a desirable standard deviation of actions caused by perturbing policy,
distance $d$ between the policy and perturbed policy is calculated as an averaged expected standard deviation and
$\sigma$ is modified to achieve certain degree of change $\delta$ in action space ($d \approx \delta$).

This method can calculate the variance of noise in parameter space that causes a certain degree of change in action space.
Therefore, an agent can conduct a certain amount of exploration despite the current policy parameters.
Additionally, this part of the covariance matrix is not biased because it considers no correlation, and all of its diagonal values are the same.
%
\subsection{Switching the two types of exploration using $\alpha$}
Finally, we describe how to switch the two types of exploration.
We propose switching the types using the difference between the acquired returns as a key to overcome problem (3):
\begin{align}
    \alpha &= \exp{\left(- h_2 \cfrac{J_{\max} - J_{\min}}{J_{\max}}  \right)}, \label{eq:im}
\end{align}
where $h_2$ is a hyper parameter, $J$ is the return, and
$J_{\max}$ and $J_{\min}$ are the maximum and minimum returns in certain episodes, respectively.
If $J_{\max}$ is negative, then $\alpha$ is calculated after $J_{\min}$ is subtracted from $J_{\max}$ and $J_{\min}$.
If $J_{\max}=0$, then we set $\alpha=1$.
$\Sigma$ becomes dominant in \eqref{eq:new_noise} when the difference is large and $\alpha$ becomes close to zero.
Directional and scalable exploration by $\Sigma$ is expected to be effective when the direction and scale can be inferred by the difference between the acquired returns.

Equation \eqref{eq:im} makes $\sigma^2 I$ dominant by setting $\alpha$ close to one when the difference between the returns is small.
The isotropic and a certain amount of exploration by $\sigma^2 I$ are expected to be effective to determine a good direction of exploration when returns have not been improved.
This part of covariance matrix also seems to be effective in a sparse reward environment to prevent the variance reduction of $\Sigma$.

Moreover, isotropic exploration is useful when a policy converges to local optima. 
In this case, returns tend to be the same.
With further exploration, an agent can escape from local optima.
Additionally, even if the variance of $\Sigma$ has reduced, then the variance of $\Sigma$ recovers because the next $\Sigma$ is calculated with the current $\sigma^2 I$ dominant noise.

In this approach, the agent activates exploration using the difference between the returns as a trigger.
Unlike intrinsic motivation, this approach does not alter the optimal solution of the policy because no exploration bonus is directly added to the reward function. 
We summarize the learning procedure in Algorithm \ref{alg:proposed_method}.
\algnewcommand\algorithmicforeach{\textbf{for each}}
\algdef{S}[FOR]{ForEach}[1]{\algorithmicforeach\ #1\ \algorithmicdo}
\begin{algorithm} 
\caption{Proposed method}
\label{alg:proposed_method}
\begin{algorithmic}
\MRepeat
    \For {episode $k \in \{1, ... K\}$}
        \State Initialize state $s$
        \State Sample noise $\varepsilon_k \sim \mathcal{N} \left(0, ~(1-\alpha)\Sigma + \alpha \sigma^2 I \right)$
        \State Perturb $\tilde{\pi} =  \pi(s;\theta + \varepsilon_k)$
        \While{episode is not done}
            \State $a_{t} \leftarrow \tilde{\pi}(s_t)$
            \State $s_{t+1}, r_{t+1} \leftarrow \text{environment}(s_{t}, a_{t})$
            \State append $(s_{t}, a_{t}, s_{t+1}, r_{t+1})$ to Replay Buffer
        \EndWhile
        \State $J_k = \sum_t r_t$
    \EndFor
    \State calculate $J_{\min}, J_{\max}$ from $\{J_k\}_{k=1}^K$
    \If {$J_{\max} < 0$} $\{J_k\}_{k=1}^K \mathrel{+}= -J_{\min}$ \EndIf
    \ForEach{$k$}
    \State $P_k = \cfrac{\exp \left(-h \cfrac{J_{\max} - J_k}{J_{\max}-J_{\min}} \right)}
                      {\sum_{k=1}^K \exp \left(-h \cfrac{J_{\max} - J_k}{J_{\max}-J_{\min}} \right)}$
    \EndFor
    \State $\Sigma \leftarrow \sum_{k=1}^K P_k \varepsilon_k \varepsilon_k^{T} $ 
    \State $\alpha = \exp{\left(- h_2 \cfrac{J_{\max} - J_{\min}}{J_{\max}}  \right)}$
    \State --------------------------------------------------------------------------
    \State \textbf{Per arbitrary cycle}
    \State Update policy $\pi(s; \theta)$ with Replay Buffer by DDPG
    \State $d(\pi, \tilde{\pi})=\frac{1}{|A|} \sum_{i=1}^{|A|}\mathbb{E}_s\left[ \left(\pi(s)_i - \tilde{\pi}(s)_i \right)^2 \right] $
    \Comment{$s \sim$ Replay Buffer}
    \State $\sigma=\begin{cases}
        1.01 \sigma & \text{if}~d(\pi, \tilde{\pi}) < \delta  \\
        \sigma /1.01 & \text{otherwise}
        \end{cases}$
\EndRepeat
\end{algorithmic}
\end{algorithm}

\section{Evaluation of a Simple Optimization Problem}\label{sec:toy_problem}
First, we evaluate the proposed method using a simple optimization problem for exploration. 
We compare the following three methods: fixed variance (FV), adaptive covariance (AC) and the proposed method (Pro).
FV, AC and Pro generate parameter space noise from $\mathcal{N}(0, \sigma_{\text{fix}}^2 I)$, $\mathcal{N}(0, \Sigma)$ and $\mathcal{N}\left(0,  (1 - \alpha)\Sigma + \alpha\sigma_{\text{fix}}^2 I\right)$ reapectively.
$\Sigma$ and $\alpha$ are calculated by \eqref{eq:cov_update} and \eqref{eq:im}.
We show the overall flow of the experiment in Algorithm \ref{alg:toy}.
\begin{algorithm} 
\caption{Toy Problem experiment}
\label{alg:toy}
\begin{algorithmic}
\If{FV} PDF = $\mathcal{N} (0, ~\sigma_{\text{fix}}^2 I)$ \EndIf
\If{AC} PDF = $\mathcal{N} (0, ~\Sigma)$ \EndIf
\If{Pro} PDF = $\mathcal{N} \left(0, ~(1-\alpha)\Sigma + \alpha \sigma_{\text{fix}}^2 I \right)$ \EndIf
\MRepeat
    \For {episode $k \in \{1, ... K\}$}
        \State Sample noise $\varepsilon_k \sim $ PDF
        \State Perturb $\tilde{\theta}_k =  \theta + \varepsilon_k$
        \State Calculate gradient $\nabla_\theta r |_{\theta = \tilde\theta_k}$
    \EndFor
    \State Calculate $r_{\min}, r_{\max}$ from $\{r_k\}_{k=1}^K$
    \If {AC or Pro}
        \ForEach{$k$}
        \State $P_k = \cfrac{\exp(-h \cfrac{r_{\max} - r_k}{r_{\max}-r_{\min}})}
                          {\sum_{k=1}^K \exp(-h \cfrac{r_{\max} - r_k}{r_{\max}-r_{\min}})}$
        \EndFor
        \State $\Sigma \leftarrow \sum_{k=1}^K P_k \varepsilon_k \varepsilon_k^{T} $ 
    \EndIf
    \If {Pro}
        \State $\alpha = \exp{\left(h_2 \cfrac{r_{\max} - r_{\min}}{r_{\max}}  \right)}$
    \EndIf
\EndRepeat
\end{algorithmic}
\end{algorithm}
Dense and sparse reward functions are defined by \begin{align}
    r_{\text{dense}} &= \exp{\left( -\|\theta - c\|^2 \right)} \\
    r_{\text{sparse}} &= \begin{cases}
                        \exp{\left( -\|\theta - c\|^2 \right)}& \mbox{if}~\|\theta - c\|^2 \leq 2.5 \\
                        0                       & \mbox{otherwise}
                        \end{cases},
\end{align}
where variable $c$ is the optimal parameter of this task.
For visualization, we parametrize a policy using two parameters: $\theta = (\theta_1, \theta_2)$.
Policy parameter $\theta$ is updated by \begin{align}
    \theta \leftarrow \theta + 0.05 \frac{1}{K} \sum_{k=1}^K \nabla_\theta r |_{\theta = \tilde\theta_k}. \label{eq:toy_update}
\end{align}
$\tilde{\theta}$ denotes the perturbed policy parameters.
At the start of each episode, $\tilde{\theta}_k$ is created by perturbing $\theta$. Then the reward acquired is calculated by $\tilde{\theta_k}$. After the $K$ episode, $\theta$ is updated by the averaged gradient of each $\tilde{\theta}_k$.
The toy problem above is conducted with a fixed 100 sets of random seeds. The initial parameter is $\theta = (0, 0)$ and the optimal parameter is set to $c=(3, 3)$. In this experiment, $\|\theta - c\| < 0.01$ is regarded as the completion of the optimization. ``Step when optimized" is calculated with the data of ``Optimized".

\subsection{Results of the Simple Optimization Problem }
The results are summarized in \tabref{tb:results}. 
Pro performed better optimization from the viewpoint of the combined speed and stability.
Pro completed optimization faster than FV.
Additionally, AC had a difficulty optimizing in a sparse reward environment. 
By contrast, Pro optimized the policy in all scenarios in these experiments.
\setlength{\tabcolsep}{8pt}
\begin{table}[t]
\centering
\caption{Results of the simple optimization problem ($K$ = 10) \newline 
Each experiment was conducted with 100 sets of fixed random seeds. 
Distance$: \|\theta - c\|$;
move: whether $\theta$ was updated from initial values; optimized: whether $\theta$ reached distance $< 0.01$. FV, AC and Pro are the abbreviations of fixed variance, adaptive covariance and the proposed method.}
\label{tb:results}
\begin{tabular}{lllll}
Dense & & & &  \\ \hline
Initial $\sigma^2$ & Step when optimized & Distance & Move & Optimized \\ \hline
FV:1.0 & 4.39e+3 $\pm$ 2.87e+2& 2.84e-4 $\pm$ 1.46e-4 & 100 & 100 \\
AC:1.0 &2.35e+3 $\pm$ 3.52e+3 & 5.06e-1 $\pm$ 1.37 & 100 & 88 \\
Pro:1.0 &1.70e+3 $\pm$ 9.09e+2 & 1.38e-4 $\pm$ 7.29e-5 & 100 & 100 \\ \hline
FV:0.5 & 2.10e+4 $\pm$ 5.87e+2& 4.43e-4 $\pm$ 2.19e-4 & 100 & 100 \\
AC:0.5 &2.45e+3 $\pm$ 3.82e+3 & 5.88e-1 $\pm$ 1.46 & 100 & 86 \\
Pro:0.5 &2.20e+3 $\pm$ 2.00e+3 & 1.04e-4 $\pm$ 4.98e-5 & 100 & 100 \\ \hline
\\
Sparse & & & &  \\ \hline
Initial $\sigma^2$ & Step when optimized & Distance & Move & Optimized \\ \hline
FV:1.0 & 4.61e+3 $\pm$ 2.78e+2& 2.79e-4 $\pm$ 1.49e-4 & 100 & 100 \\
AC:1.0 &1.14e+3 $\pm$ 1.24e+2 & 2.34 $\pm$ 2.08 & 64 & 44 \\
Pro:1.0 &1.25e+3 $\pm$ 9.77e+1 & 1.38e-4 $\pm$ 6.76e-5 & 100 & 100 \\ \hline
FV:0.5 & 2.81e+4 $\pm$ 8.77e+2& 4.13e-2 $\pm$ 3.13e-1 & 100 & 96 \\
AC:0.5 &1.17e+3 $\pm$ 1.68e+2 & 3.08 $\pm$ 1.88 & 36 & 27 \\
Pro:0.5 &4.25e+3 $\pm$ 1.39e+3 & 1.09e-4 $\pm$ 6.16e-5 & 100 & 100 \\ \hline
\end{tabular}
\end{table}

\begin{figure}[t]
    \centering
    \includegraphics[width=122mm]{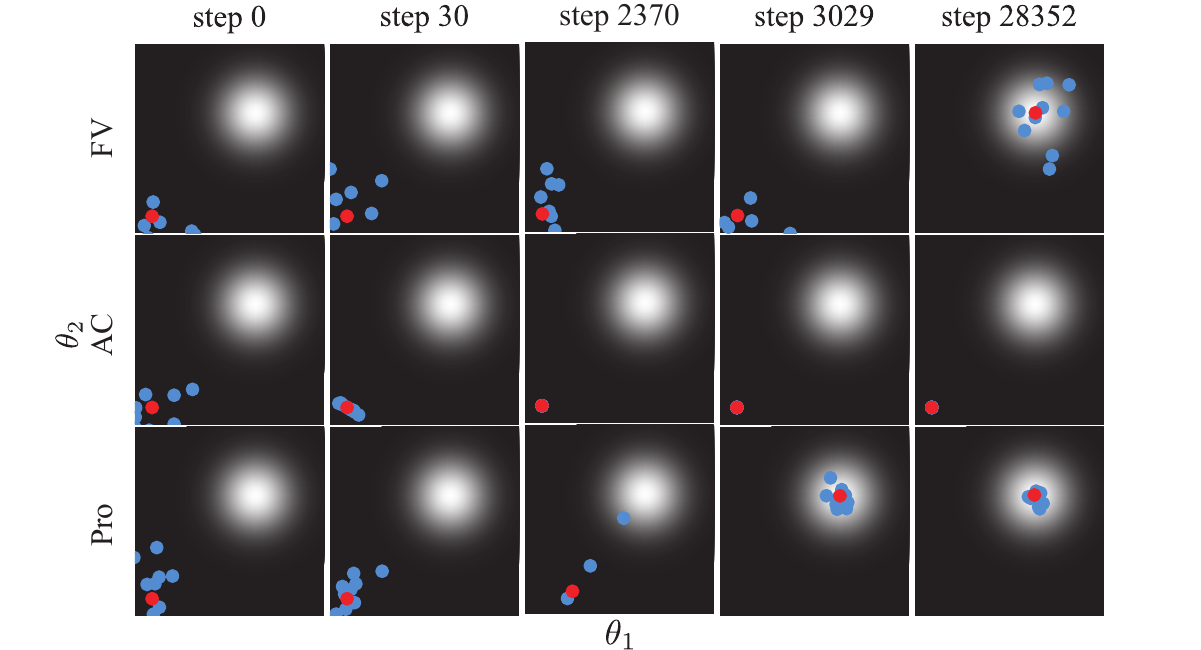}
    \caption{Optimization process with the sparse reward function \newline
Red dots signify current policy parameter $\theta$;
blue dots signify perturbed parameters $\theta_k$ by injecting noise from each distribution;
the background color corresponds to the reward; a white area corresponds to a higher reward.
FV, AC and Pro are the abbreviations of fixed variance, adaptive covariance and the proposed method. }
    \label{fig:toy_pro}
\end{figure}

Additionally, the optimization processes are visualized in \figref{fig:toy_pro}.
FV took more time to optimize than Pro; however, it completed the optimization.
AC could not reach the optimal parameter for the random seeds.
As mentioned in subsection \ref{subsec:reason}, the variance of parameter space noise decreased sharply and could not conduct the exploration.
Pro did not reduce the variance of the distribution.
When it acquired rewards, it generated directional noise toward the axis of the rewards. 
This method achieved much faster optimization than FV.
\section{Experiments}
We evaluated the proposed method for baseline tasks of deep RL and a sparse reward environment.
We compared it with Plappert {\it et al}.,'s method \cite{plappert2017parameter}.
%
\subsection{Experimental setup of the baseline tasks}
The proposed approach was tested on the continuous control tasks from OpenAI-Gym \cite{brockman2016openai} using the MuJoCo \cite{todorov2012mujoco} physics simulator.
The architecture of the neural networks and hyper-parameters was almost same as previous work \cite{plappert2017parameter}. Both the actor and critic consisted of two hidden layers, which each had 64 neurons and the ReLU activation function. 
The output layer of the actor had tanh as an activation function. 
The actions were calculated by multiplying the torque limit and actor outputs.
The critic received actor tanh outputs at its second hidden layer.
Input states were normalized by estimating their mean and variance online. 
Layer normalization \cite{ba2016layer} was applied to all hidden layers.
The replay buffer contained a transition of up to 1,000K states.
The actor and critic were soft updated at $\tau=0.01$ with the target network.
The learning rate of the actor was $10^{-4}$ and that of the critic was $10^{-3}$.
Discount rate $\gamma=0.99$ was used.
The Adam optimizer \cite{kingma2015adam} was applied to optimize both the actor and critic with batch size 64.
The critic was regularized using an $L2$ penalty with coefficient $10^{-2}$. 
The initial variance of parameter space noise $\sigma^2$ was set to $\sigma=0.2$ in the dense reward environment and $\sigma=0.6$ in the sparse reward environment for the prior method.
The hyper parameters of the proposed method were as follows: $h=8.0$ in \eqref{eq:weight}, number of episodes to update the covariance $K=10$ in \eqref{eq:cov_update} \eqref{eq:im} and $h_2=10.0$ in \eqref{eq:im}.
\subsection{Results of the baseline tasks}
\begin{figure}[t]
  \centering
  \includegraphics[width=110mm]{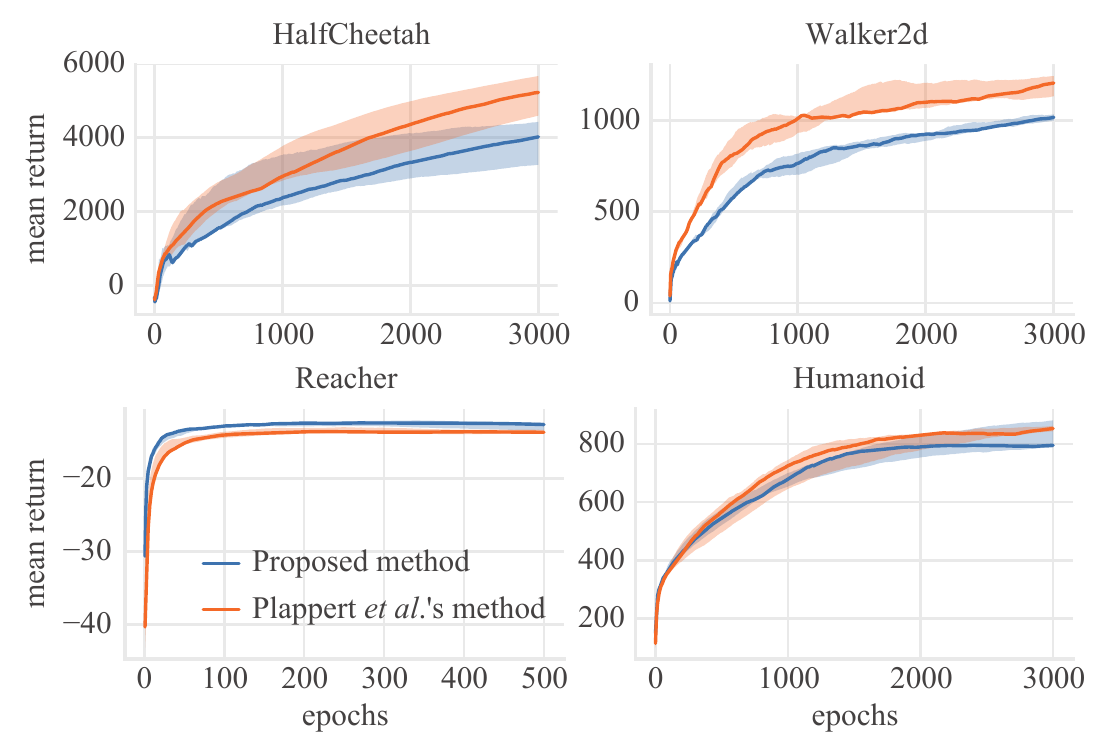}
  \caption{Learning results of the baseline tasks \newline The colored lines and shading correspond to medians and interquartile ranges of the averaged return in each epoch. The experiment for each environment was conducted in five sets of fixed random seeds. The return was acquired using the perturbed policy.}
  \label{fig:results_dense}
\end{figure}

The results are shown in \figref{fig:results_dense}.
The proposed method achieved competitive performances compared with the previous method. 
These results suggest that the agent explored sufficiently with both directional and non-directional noise in dense reward environments. 
\subsection{Experiments for sparse reward environments}
The proposed method was also tested in continuous control tasks with sparse reward environments.
We used three sparse reward environments: SparseCartpoleSwingup, SparseDoublePendlumSwinup and SparseHalfCheetah. These environments were created by modifying environments in rllab \cite{duan2016benchmarking}, and were used in previous works for evaluation \cite{plappert2017parameter,houthooft2016vime}.
In the SparseCartpoleSwingup task, an agent received reward $+1$ only when the agent swung up the pole within a specific range. 
SparseDoublePendlum only yielded a reward if the agent reached the upright position.
SparseHalfCheetah only yielded reward +1 when the agent was beyond 5 m.
All the tasks had the same time horizon of 500 steps.
\subsection{Results for sparse reward environments}
\begin{figure}[t]
  \centering
  \includegraphics[width=110mm]{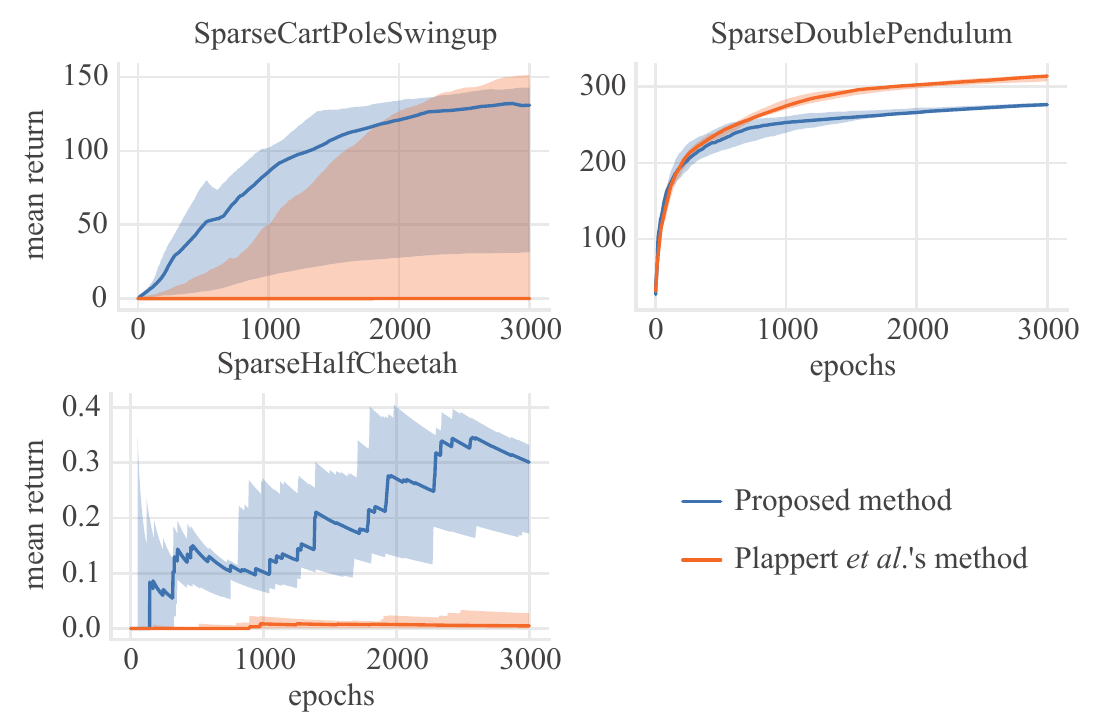}
  \caption{Learning results in sparse reward tasks \newline The colored lines and shading correspond to medians and interquartile ranges of the averaged return in each epoch. The experiment for each environment was conducted in 10 sets of fixed random seeds. The return was acquired using the perturbed policy.}
  \label{fig:results_sparse}
\end{figure}

The results are summarized in \figref{fig:results_sparse}.
The proposed method achieved better median performance in SparseCartPoleSwingup than the previous method.
In SparseDoublePendlum, it obtained a competitive policy.
In SparseHalfCheetah, our approach showed better results overall.
In the sparse reward environment, it was essential to acquire a reward continuously.
If adaptive covariance acquired the reward, then it led an agent to explore in the direction of the reward.
Although the previous method also acquired a reward with a small probability,
the ratio to transient states with non-zero rewards was small.
Therefore, the previous method had a difficulty in optimizing policy.
The results of SparseDoublePendlum were not substantially different because this environment did not seem to require exploration. It was shown in \cite{plappert2017parameter} that agents without exploration can learn the policy.
\subsection{Analysis of exploration}
\begin{figure}[t]
  \centering
  \includegraphics[width=122mm]{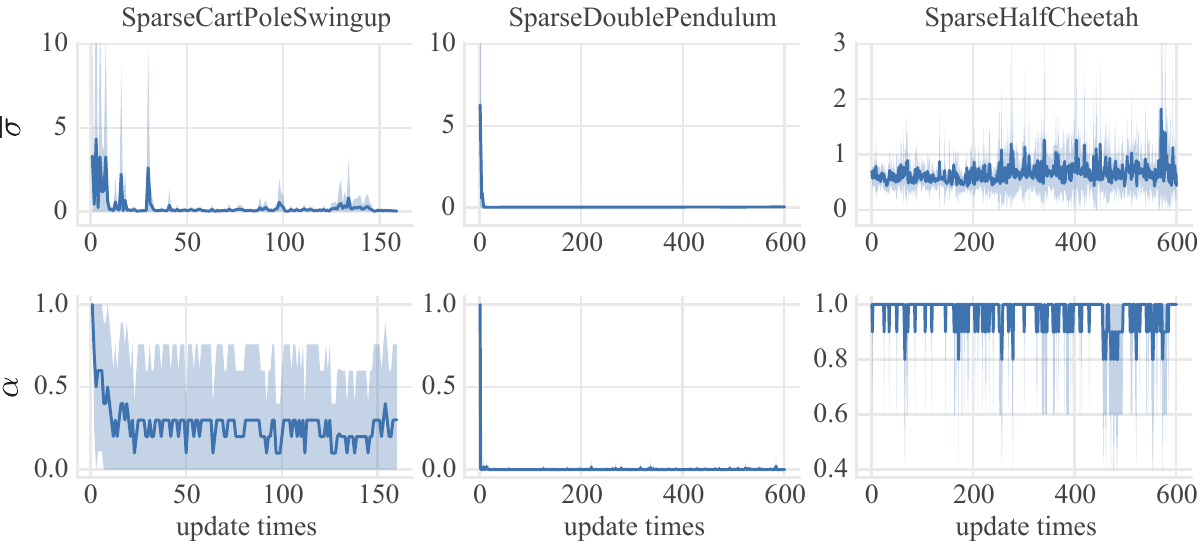}
  \caption{Analysis of exploration in sparse reward environments \newline The horizontal axis corresponds to the number of updating $\Sigma$. The colored lines and shading correspond to means and standard deviations in 10 agents. We calculate them from the start to the minimum ``update times" because it changes according to how to reset an episode. We display values clipped by the domain of each variables.}
  \label{fig:stdev_sparse}
\end{figure}
We analyze how the proposed method managed exploration in the sparse reward environment.
We analyze two variables relating to exploration.
One is the variance of the noise distribution.
The variance indicates the degree of exploration because it controls the power of noise.
We evaluate only a part of the covariance matrix $\Sigma$.
It reflects the feature of current noise using \eqref{eq:cov_update} whichever part of the covariance matrix is dominant.
We calculate the trend of the standard deviation of noise distribution using
\begin{align}
    \overline{\sigma} = \sqrt{\frac{1}{N} \sum_{i=1}^N \Sigma_{ii}},
\end{align}
where $N$ is the number of policy parameters.

The other variable is $\alpha$ in \eqref{eq:im}. This parameter switches exploration types by modifying the dominance between $\sigma^2 I$ and $\Sigma$. $\alpha\!=\!1$ corresponds to an isotropic exploration. $\alpha\!=\!0$ corresponds to directional and scaled exploration.

The results are shown in \figref{fig:stdev_sparse}.
In SparseCartPoleSwingup, isotropic exploration was dominant in the first stage of learning.
The scale of noise was large at the same stage. 
We believe the reason is that the policy parameters were still far from the optimal policy, and more perturbation led to better performance.
After the middle stage of the learning process, $\overline{\sigma}$ became small.
It is expected that the policy matured, and a smaller perturbation led to a higher return.
In SprseDoublePendulum, $\overline{\sigma}$ had a tiny value and $\alpha\!=\!0$, except for the first few iterations.
This environment did not seem to require exploration because it was shown in \cite{plappert2017parameter} that agents without exploration can learn the policy.
In the SparseHalfCheetah environment, agents performed more exploratory behaviors. $\overline{\sigma}$ had a larger value overall. 
This environment may have required exploration because it had larger state-action space than the other two environments. 
In some phases, $\alpha$ had a small value and exploited the adaptivity of $\Sigma$. 
This may have enabled an agent to efficiently gather information that contained a high return and learn faster.
\section{Discussion}
We experimentally demonstrated competitiveness in a dense reward environment and an advantage in a sparse reward environment by comparing the proposed method with a previous method. 
The proposed method managed the correlation of parameter space noise, which was not used in previous works \cite{plappert2017parameter,fortunato2017noisy}.
Our approach continuously switched isotropic exploration and directional exploration. Our switching method can be a solution to the problem that remained in the previous work \cite{colas2018gep}.

To investigate the performance at convergence is important.
In this paper, we do not show that our approach acquired a better final policy than the previous method because we terminated learning halfway because of time constraints.
Our approach can be trapped in local optima because it perturbs the policy along the axis, where locally higher rewards are observed.
However, it is possible that our approach can eventually escape local optima.
Equation \eqref{eq:im} activates exploration when the return has converged in addition to when an agent cannot obtain any return.

The limitation of our approach is a deceptive reward environment.
The maze task with a rat guard is one of the examples.
The proposed method cannot escape from such a trap because it uses a reward signal directly to determine the exploration strategy.
In this environment, intrinsic motivation with an exploration bonus performs better.

The computational complexity is another problem.
The larger the network, the faster the increase in the amount of noise to sample because noise is needed for all weights and biases.
In this case, the Local Reparameterization Trick \cite{kingma2015variational} is one of the solutions.
This method can reduce the amount of noise to the number of all neurons.
\section{Conclusions}
In this paper, we presented a new parameter space noise method for efficient exploration in reinforcement learning. The proposed method switched isotropic exploration and directional exploration using the differences of accumulated returns as a key.  The direction is determined by the correlations and scales of noises which have led to higher returns in previous episodes.

Experiments in sparse rewards environments verified that our method could learn better policy in those environments than Plappert {\it et al}.'s method \cite{plappert2017parameter}. 
The proposed method could proceed stable learning in the SparseCartpoleSwingup and SparseHalfCheetah environment which the previous method was difficult to learn. Our experiments also demonstrated that the proposed method did not improve the performance in the dense reward environment, though it could acquire competitive performance with the prior method.

Our analysis of exploration showed that our approach first executed exploration with isotropic and large-scale noise and then switched exploration type to directional and small-scale noise in SparseCartPoleSwingup and SparseDoublePendulum.
In SparseHalfCheetah, although the noise had large values, an agent used directional exploration in some phases.
These results suggest that biasing exploration is important especially in sparse reward environment to continuously obtain rewards and learn policy. 

Our proposed method is one of the approaches to modify the strategy of exploration automatically.
Although we switched exploration just using return,  more structured exploration like using Bayesian, which is closely related to parameter space noise, is future work. 

\section*{Acknowledgements}
This research was supported by the JSPS under a Grant-in-Aid for Scientific Research
(A) JP26240039 and a Grant-in-Aid for Young Scientists (A) JP15H05320.
We would like to thank OpenAI for providing useful implementations of deep RL in the Baselines repository \cite{baselines}

%
\bibliographystyle{splncs04}
\bibliography{bib/RL}

\begin{thebibliography}{10}
\providecommand{\url}[1]{\texttt{#1}}
\providecommand{\urlprefix}{URL }
\providecommand{\doi}[1]{https://doi.org/#1}

\bibitem{auer2002finite}
Auer, P., Cesa-Bianchi, N., Fischer, P.: Finite-time analysis of the multiarmed
  bandit problem. Machine learning  \textbf{47}(2-3),  235--256 (2002)

\bibitem{ba2016layer}
Ba, J.L., Kiros, J.R., Hinton, G.E.: Layer normalization. arXiv preprint
  arXiv:1607.06450  (2016)

\bibitem{bellemare2016unifying}
Bellemare, M., Srinivasan, S., Ostrovski, G., Schaul, T., Saxton, D., Munos,
  R.: Unifying count-based exploration and intrinsic motivation. In: Advances
  in Neural Information Processing Systems. pp. 1471--1479 (2016)

\bibitem{brockman2016openai}
Brockman, G., Cheung, V., Pettersson, L., Schneider, J., Schulman, J., Tang,
  J., Zaremba, W.: Openai gym. arXiv preprint arXiv:1606.01540  (2016)

\bibitem{colas2018gep}
Colas, C., Sigaud, O., Oudeyer, P.Y.: Gep-pg: Decoupling exploration and
  exploitation in deep reinforcement learning algorithms. arXiv preprint
  arXiv:1802.05054  (2018)

\bibitem{baselines}
Dhariwal, P., Hesse, C., Klimov, O., Nichol, A., Plappert, M., Radford, A.,
  Schulman, J., Sidor, S., Wu, Y.: Openai baselines.
  \url{https://github.com/openai/baselines} (2017)

\bibitem{duan2016benchmarking}
Duan, Y., Chen, X., Houthooft, R., Schulman, J., Abbeel, P.: Benchmarking deep
  reinforcement learning for continuous control. In: International Conference
  on Machine Learning. pp. 1329--1338 (2016)

\bibitem{fortunato2017noisy}
Fortunato, M., Azar, M.G., Piot, B., Menick, J., Osband, I., Graves, A., Mnih,
  V., Munos, R., Hassabis, D., Pietquin, O., et~al.: Noisy networks for
  exploration. arXiv preprint arXiv:1706.10295  (2017)

\bibitem{hansen1996adapting}
Hansen, N., Ostermeier, A.: Adapting arbitrary normal mutation distributions in
  evolution strategies: the covariance matrix adaptation. In: Proceedings of
  IEEE International Conference on Evolutionary Computation. pp. 312--317
  (1996)

\bibitem{houthooft2016vime}
Houthooft, R., Chen, X., Duan, Y., Schulman, J., De~Turck, F., Abbeel, P.:
  Vime: Variational information maximizing exploration. In: Advances in Neural
  Information Processing Systems. pp. 1109--1117 (2016)

\bibitem{kingma2015adam}
Kingma, D.P., Ba, J.: Adam: A method for stochastic optimization. In:
  International Conference on Learning Representations (2015)

\bibitem{kingma2015variational}
Kingma, D.P., Salimans, T., Welling, M.: Variational dropout and the local
  reparameterization trick. In: Advances in Neural Information Processing
  Systems. pp. 2575--2583 (2015)

\bibitem{lehman2017more}
Lehman, J., Chen, J., Clune, J., Stanley, K.O.: Es is more than just a
  traditional finite-difference approximator. arXiv preprint arXiv:1712.06568
  (2017)

\bibitem{lillicrap2015continuous}
Lillicrap, T.P., Hunt, J.J., Pritzel, A., Heess, N., Erez, T., Tassa, Y.,
  Silver, D., Wierstra, D.: Continuous control with deep reinforcement
  learning. arXiv preprint arXiv:1509.02971  (2015)

\bibitem{mnih2016asynchronous}
Mnih, V., Badia, A.P., Mirza, M., Graves, A., Lillicrap, T., Harley, T.,
  Silver, D., Kavukcuoglu, K.: Asynchronous methods for deep reinforcement
  learning. In: International Conference on Machine Learning. pp. 1928--1937
  (2016)

\bibitem{mnih2015human}
Mnih, V., Kavukcuoglu, K., Silver, D., Rusu, A.A., Veness, J., Bellemare, M.G.,
  Graves, A., Riedmiller, M., Fidjeland, A.K., Ostrovski, G., et~al.:
  Human-level control through deep reinforcement learning. Nature
  \textbf{518}(7540),  529--533 (2015)

\bibitem{pathak2017curiosity}
Pathak, D., Agrawal, P., Efros, A.A., Darrell, T.: Curiosity-driven exploration
  by self-supervised prediction. In: International Conference on Machine
  Learning (2017)

\bibitem{peters2005natural}
Peters, J., Vijayakumar, S., Schaal, S.: Natural actor-critic. In: European
  Conference on Machine Learning. pp. 280--291. Springer (2005)

\bibitem{plappert2017parameter}
Plappert, M., Houthooft, R., Dhariwal, P., Sidor, S., Chen, R.Y., Chen, X.,
  Asfour, T., Abbeel, P., Andrychowicz, M.: Parameter space noise for
  exploration. arXiv preprint arXiv:1706.01905  (2017)

\bibitem{ruckstiess2008state}
R{\"u}ckstie{\ss}, T., Felder, M., Schmidhuber, J.: State-dependent exploration
  for policy gradient methods. Machine Learning and Knowledge Discovery in
  Databases pp. 234--249 (2008)

\bibitem{salimans2017evolution}
Salimans, T., Ho, J., Chen, X., Sutskever, I.: Evolution strategies as a
  scalable alternative to reinforcement learning. arXiv preprint
  arXiv:1703.03864  (2017)

\bibitem{schulman2015trust}
Schulman, J., Levine, S., Abbeel, P., Jordan, M., Moritz, P.: Trust region
  policy optimization. In: International Conference on Machine Learning. pp.
  1889--1897 (2015)

\bibitem{silver2014deterministic}
Silver, D., Lever, G., Heess, N., Degris, T., Wierstra, D., Riedmiller, M.:
  Deterministic policy gradient algorithms. In: International Conference on
  Machine Learning (2014)

\bibitem{stulp2013robot}
Stulp, F., Sigaud, O.: Robot skill learning: From reinforcement learning to
  evolution strategies. Paladyn, Journal of Behavioral Robotics  \textbf{4}(1),
   49--61 (2013)

\bibitem{such2017deep}
Such, F.P., Madhavan, V., Conti, E., Lehman, J., Stanley, K.O., Clune, J.: Deep
  neuroevolution: Genetic algorithms are a competitive alternative for training
  deep neural networks for reinforcement learning. arXiv preprint
  arXiv:1712.06567  (2017)

\bibitem{sukhbaatar2017intrinsic}
Sukhbaatar, S., Kostrikov, I., Szlam, A., Fergus, R.: Intrinsic motivation and
  automatic curricula via asymmetric self-play. arXiv preprint arXiv:1703.05407
   (2017)

\bibitem{todorov2012mujoco}
Todorov, E., Erez, T., Tassa, Y.: Mujoco: A physics engine for model-based
  control. In: IEEE/RSJ International Conference on Intelligent Robots and
  Systems. pp. 5026--5033 (2012)

\bibitem{williams1992simple}
Williams, R.J.: Simple statistical gradient-following algorithms for
  connectionist reinforcement learning. Machine learning  \textbf{8}(3-4),
  229--256 (1992)

\bibitem{zhang2017relationship}
Zhang, X., Clune, J., Stanley, K.O.: On the relationship between the openai
  evolution strategy and stochastic gradient descent. arXiv preprint
  arXiv:1712.06564  (2017)

\end{thebibliography}
\end{document}